\documentclass[runningheads]{llncs}
\usepackage[T1]{fontenc}
\usepackage{graphicx}
\usepackage{hyperref}
\usepackage{multirow}
\begin{document}
\title{Improved Multi-Task Brain Tumour Segmentation with Synthetic Data Augmentation}
\titlerunning{Improved Segmentation with Synthetic Data}
%
\author{André Ferreira \inst{1,2,3,10,11}\orcidID{0000-0002-9332-0091} \and
Tiago Jesus \inst{1,4,5}\orcidID{0000-0003-1437-5439} \and  
Behrus Puladi \inst{10,11}\orcidID{0000-0001-5909-6105}  \and
Jens Kleesiek\inst{3,6,7,12} \and
Victor Alves\orcidID{0000-0003-1819-7051}\inst{1} \and
Jan Egger\inst{2,3,6,8}
}
\authorrunning{A. Ferreira et al.}
%
\institute{Center Algoritmi / LASI, University of Minho, Braga, 4710-057,  Portugal 
\and
Computer Algorithms for Medicine Laboratory, Graz, Austria
\and
Institute for AI in Medicine (IKIM), University Medicine Essen, Girardetstraße 2, Essen, 45131, Germany
\and Life and Health Sciences Research Institute (ICVS), School of Medicine, University of Minho, Braga, Portugal
\and ICVS/3B’s – PT Government Associate Laboratory, Braga, Guimarães, Portugal
\and Cancer Research Center Cologne Essen (CCCE), University Medicine Essen, Hufelandstraße 55, Essen, 45147, Germany
\and German Cancer Consortium (DKTK), Partner Site Essen, Hufelandstraße 55, Essen, 45147, Germany
\and Institute of Computer Graphics and Vision, Graz University of Technology, Inffeldgasse 16, Graz, 8010, Austria
\and Department of Neurosurgery and Spine Surgery, University Hospital Essen, Essen, Germany\\
\and Institute of Medical Informatics, University Hospital RWTH Aachen, Aachen, Germany
\\
\and Department of Oral and Maxillofacial Surgery, University Hospital RWTH Aachen, Aachen,  Germany \\
\and Department of Physics, TU Dortmund University, Dortmund, Germany
\email{\{id10656\}@alunos.uminho.pt}}
\maketitle              

\begin{abstract}

This paper presents the winning solution of task 1 and the third-placed solution of task 3 of the BraTS challenge. The use of automated tools in clinical practice has increased due to the development of more and more sophisticated and reliable algorithms. However, achieving clinical standards and developing tools for real-life scenarios is a major challenge. To this end, BraTS has organised tasks to find the most advanced solutions for specific purposes. In this paper, we propose the use of synthetic data to train state-of-the-art frameworks in order to improve the segmentation of adult gliomas in a post-treatment scenario, and the segmentation of meningioma for radiotherapy planning. Our results suggest that the use of synthetic data leads to more robust algorithms, although the synthetic data generation pipeline is not directly suited to the meningioma task. In task 1, we achieved a DSC of 0.7900, 0.8076, 0.7760, 0.8926, 0.7874, 0.8938 and a HD95 of 35.63, 30.35, 44.58, 16.87, 38.19, 17.95 for ET, NETC, RC, SNFH, TC and WT, respectively and, in task 3, we achieved a DSC of 0.801 and HD95 of 38.26, in the testing phase. The code for these tasks is available at \href{https://github.com/ShadowTwin41/BraTS_2023_2024_solutions}{https://github.com/ShadowTwin41/BraTS\_2023\_2024\_solutions}.

\end{abstract}
\section{Introduction}
Gliomas are a type of brain tumour originating from glial cells \cite{agosti2024,ferreira2024we}. This type of tumour is among the deadliest types of cancer and constitutes the most prevalent malignant primary brain tumours in adults \cite{de20242024}. Gliomas' aggressive nature and resistance to therapy make them a major problem in oncology \cite{agosti2024}.
As such, promptly and correctly identifying them is crucial for an effective treatment and post-treatment. Therefore, the development of algorithms to automatically detect and segment the gliomas, as the Brain Tumour Segmentation (BraTS) challenges propose, would help patients worldwide.

Magnetic resonance imaging (MRI) is a type of medical imaging modality that remains the gold standard imaging method for post-treatment across the spectrum of gliomas. The patients' MRI exams provide crucial information on tumour size, location, and morphological changes over time. Post-treatment imaging of gliomas is fundamental and significantly influences clinical decision-making and outcomes on patients \cite{de20242024}.

Meningioma is another type of brain tumour, comprising 40.8\% of all central nervous system (CNS) tumours and 55.4\% of all non-malignant CNS tumours, making it the most common primary intracranial tumour.
The vast majority, 99.1\%, of meningiomas, are non-malignant and can be followed with MRI exams if asymptomatic \cite{labella2024brain,labella2023asnr,labella2024analysis}. Although segmentation of pre-operative meningioma is essential for patients' treatment, it is a complex and very time-consuming task. As such, just like the gliomas' task, the BraTS challenge meningioma task aims to create a community benchmark for automated segmentation of these tumours which will save time and improve patients' radiotherapy planning.

\subsection{State of the art}
In 2024, the first task will switch from the segmentation of adult gliomas to post-treatment of adult gliomas. Even though these tasks have specific characteristics and challenges, the tools used for the first can be applied and fine-tuned for the new task. 

Since the first edition of the BraTS challenge, multiple different approaches have been developed. Deep convolutional networks have performed the best since 2014 \cite{pereira2016brain,isensee2021nnu}. The nnUNet \cite{isensee2021nnu_nnunet}
has dominated the competition since 2020. The recent winners have also integrated the nnUNet pipeline into their solution, as the U-Net offers a large capacity for segmentation and the robustness of the nnUNet. 

In the 2020 edition, \cite{isensee2021nnu} introduced the nnUNet. Some specific changes were made to the basic nnUNet pipeline: Use of larger batch size (from 2 to 5), use of more aggressive data augmentation, use of batch normalisation, use of batch dice, and threshold-based post-processing. In 2021, \cite{luu2021extending} extended the nnUNet pipeline by increasing the size of the encoder, replacing batch normalisation with group normalisation and preserving batch size 2. In 2022 \cite{zeineldin2022multimodal} the ensemble of three architectures was used: DeepScan \cite{mckinley2019ensembles}, DeepSeg \cite{zeineldin2020deepseg}, and the nnUNet developed by the winner of 2020 edition. In 2023, the winning solution \cite{ferreira2024we} uses an ensemble of three networks using two methods of data augmentation for training. The networks (nnUNet\cite{isensee2021nnu_nnunet}, Swin\cite{hatamizadeh2021swin} and nnUNet of the 2021 winner \cite{luu2021extending}) were trained with the pipeline provided by the nnUNet. Data augmentation with GANs and registration was used to increase the amount of data available for training.

The segmentation of meningiomas in radiotherapy (Task 3) is similar to the segmentation of gliomas in adults but differs from the type of tumour. Meningiomas are usually benign extra-axial tumours with different radiological and anatomical appearances and a tendency to multiplicity. The dataset used last year is very different from the one used this year, as last year there were four modalities (T2, T2/FLAIR, T1, and T1Gd), whereas this year only one modality (pre-radiation therapy planning brain MRI T1Gd) is present in the dataset. The segmentation has also changed, as this year only the gross tumour volume (GTV) was segmented. Although the datasets have differences, the tools used may be similar. In 2023, the winning team used a similar tool to nnUNet, the \textit{Auto3dseg} \cite{myronenko2023auto3dseg}, which was developed by Nvidia and is available in the MONAI library. Once again, a self-configuring U-Net-based solution shown very strong performance.  

Our approach this year (2024) aims to compare the default 3D full resolution nnUNet with the MedNeXt \cite{roy2023mednext}, as the latter provided better results on the adult glioma segmentation task \cite{isensee2024nnu_revisited}, but there is no study on the effectiveness of this network on the adult glioma post-treatment and meningioma radiotherapy tasks. We also applied the same data augmentation with GANs used by \cite{ferreira2024we}.

\section{Methods}

Two machines were used for the experiments. The first machine is an IKIM cluster node with 6 NVIDIA RTX 6000, 48 GB of VRAM, 1024 GB of RAM, and AMD EPYC 7402 24-Core Processor. The second is a RWTH aachen cluster with NVIDIA H100 GPUs with 96 GB of VRAM, 512GB of RAM and CPUs Intel Xeon 8468 Sapphire.  Only one GPU was used per training.

\subsection{Datasets}

\subsubsection{Task 1 - Adult Glioma Post Treatment:}
The dataset is composed of 2200 samples (training (70\%), validation (10\%), and testing datasets (20\%)) containing all 4 modalities: native (T1), post-contrast T1-weighted (T1Gd), T2-weighted (T2), and T2 Fluid Attenuated Inversion Recovery (FLAIR). These volumes were acquired with different clinical protocols and various scanners from multiple data-contributing institutions. All data was co-registered to the same anatomical template, interpolated to the same resolution (1 $mm^3$), and skull stripped. All cases were manually annotated. The annotations contain the non-enhancing tumour core (NETC — label 1), the surrounding non-enhancing FLAIR hyperintensity (SNFH) — label 2), the enhancing tissue (ET — label 3), and the resection cavity (RC - label 4) \cite{de20242024}.

\subsubsection{Task 3 - Meningioma Radiotherapy:}
The dataset was changed from 2023's edition in order to be more clinically relevant \cite{labella2024brain,labella2023asnr,labella2024analysis}. The dataset consists of 790 radiotherapy planning brain MRI exams (500 for training, 70 for validation and 220 for testing). Only post-contrast T1-weighted (T1Gd) scans and gross tumour volume (GTV) segmentations in the native acquisition space are available. No skull stripping or co-registration was made, only defacing to preserve patient anonymity.

\subsection{Data augmentation} \label{DataAugmentation}
\begin{figure}[h!]
     \centering
     \includegraphics[width=\textwidth]{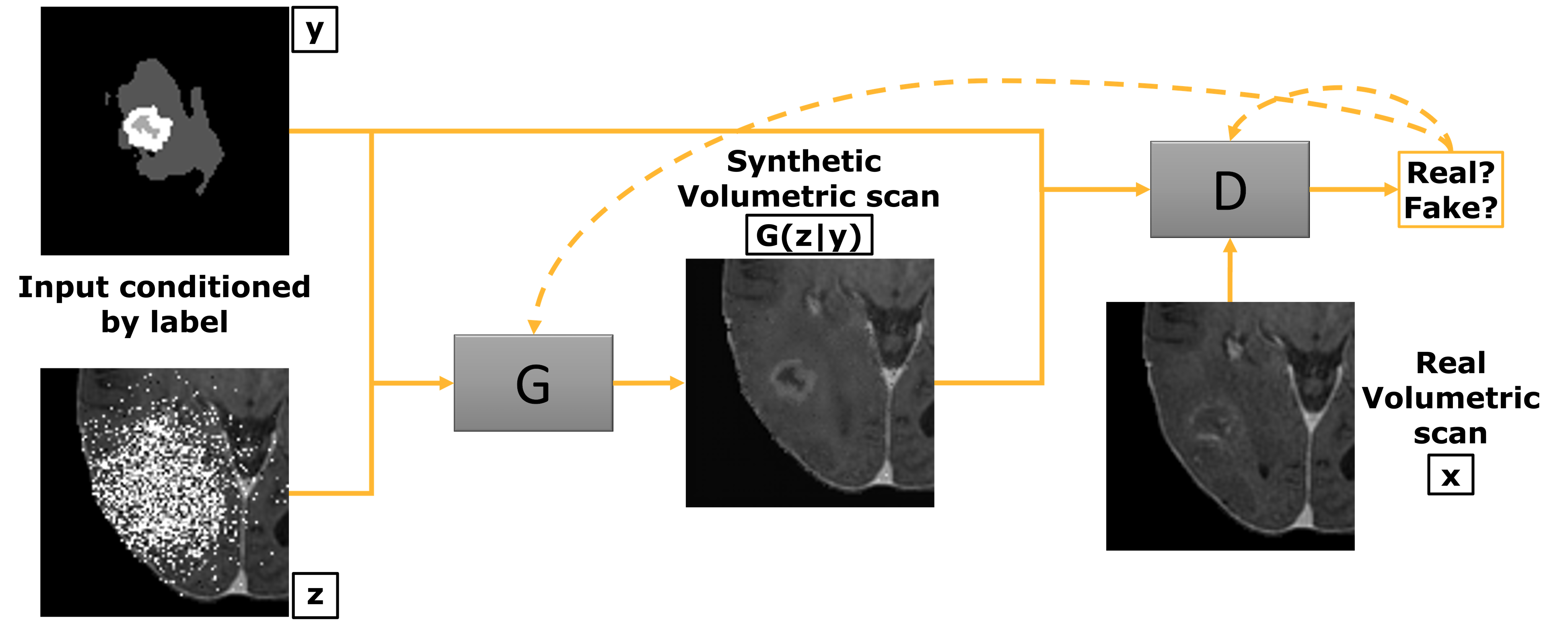}
        \caption{The training pipeline of the \textbf{GliGAN}. The noise scan (z) and the label (y) are concatenated and fed into the generator (G). The discriminator (D) assesses the realism of a real scan and the reconstruction. Figure taken from \cite{ferreira2024we}.}
        \label{fig:GANs-train}
\end{figure}

The same approach as in \cite{ferreira2024we} was used to train two different GliGANs for both Task 1 and Task 3 in order to insert synthetic tumours into the healthy part of the scans, as shown in Figure \ref{fig:GANs-train}. This strategy aims to increase the variability of tumours and reduce the class imbalance between healthy and unhealthy tissue. A random label generator based on \cite{ferreira2022generation} was also trained to generate random labels in both tasks and to be used in the trained GliGANs. We randomly choose whether to use a real label or a generated label for inference.

This strategy expands the training data, which leads to a higher variability of the training data, likely increasing the robustness of the trained models and reducing the number of false negatives (FN). A more detailed explanation can be found in \cite{ferreira2024we}. For Task 1, 14,301 new cases were created, resulting in a total of 15,651 cases for training (14,301 + 1,350). For Task 3, 6,970 new cases were created, resulting in 7,470 cases (6,970+500).

It is important to note that each GAN was trained individually, i.e. no data from one dataset was used to train the GAN of the other dataset. Both tasks were developed independently of each other. To ensure the realism of the synthetic tumours, visual inspections were performed at several checkpoints. The loss graphs were also observed to ensure that the training of the GAN did not collapse. We observed very stable training in all situations. 

\begin{figure}
\includegraphics[width=\textwidth]{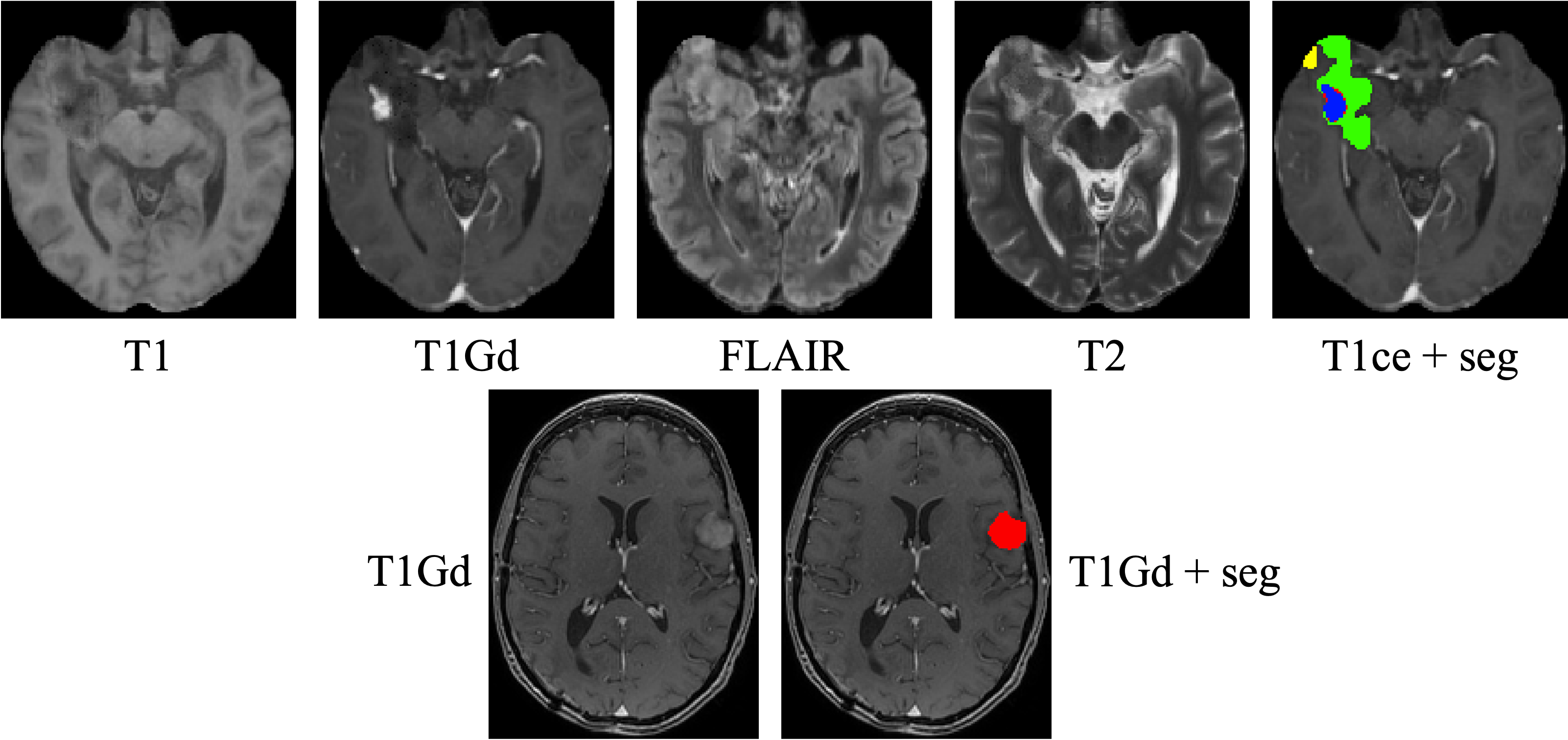}
\caption{First row: Sample 00005-100 of the training set of Task 1 with a synthetic tumour in the healthy part of the brain in all 4 modalities and the corresponding segmentation. Second row: sample 0002-1 of the training set of Task 3 with a synthetic tumour in the healthy part of the brain and the corresponding segmentation.} \label{fig:fake_samples}
\end{figure}

Only minimal changes have been made to the pipeline for training and generating synthetic data, but these should be emphasised:
For Task 1, the number of channels used as input was changed from 4 to 5 due to the introduction of the new region (RC);
For Task 3, the number of input channels was also changed from 4 to 2, as only the GTV is used. Since the scans were not pre-processed and skull-stripped in task 3, the pipeline is not able to detect where the skull is located, so it is possible that some tumours were created in the skull. In addition, Otsu's thresholding technique was used to identify the foreground and background in order to place the tumour in the head. Figure \ref{fig:fake_samples} shows an example where a synthetic tumour was inserted into a case of Task 1 and Task 3 datasets.

\subsection{Networks}
\begin{figure}
\includegraphics[width=\textwidth]{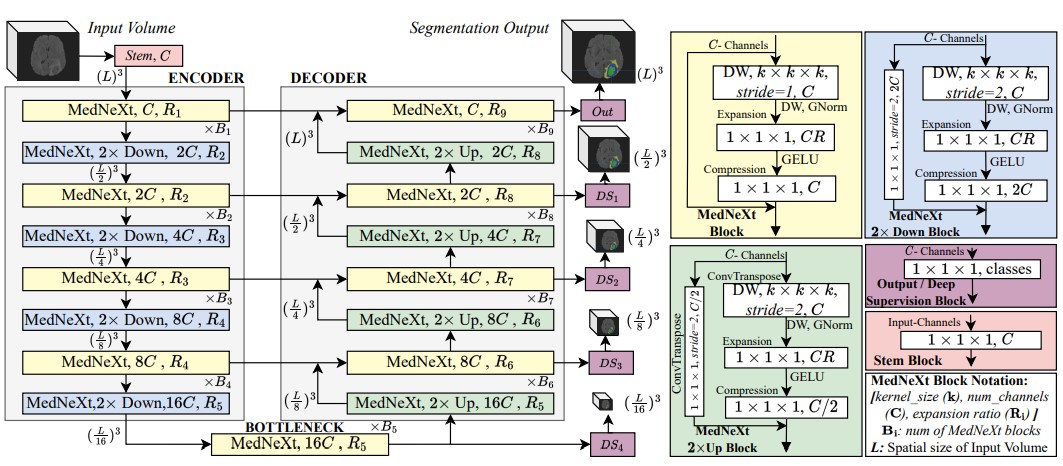}
\caption{Architectural design of the MedNeXt \cite{roy2023mednext}.} \label{fig:mednext_architecture}
\end{figure}

Several network architectures were experimented with in order to ascertain which of them attained the best segmentation results. These networks were all based on U-nets, a type of architecture well proven to show good results in brain segmentation \cite{ferreira2024we,jesus2024}. One of the architectures experimented with was the MedNeXt, depicted in Figure \ref{fig:mednext_architecture} which uses the nnUnet pipeline. This architecture is convolutional and retains the inductive bias inherent to ConvNets, allowing for easier training on sparse medical datasets \cite{roy2023mednext}. From the several network sizes offered by MedNeXt the L size network with 5x5x5 kernel was used as it could better handle the complexity of the BraTS challenge dataset exams.

\begin{figure}
\includegraphics[width=\textwidth]{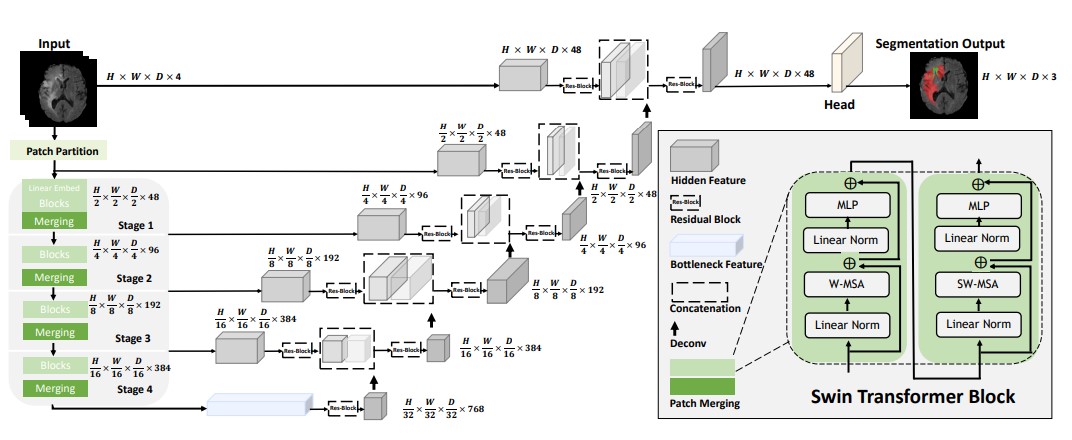}
\caption{Swin UNETR's architecture overview \cite{hatamizadeh2021swin}.} \label{fig:swinunetr_architecture}
\end{figure}

Another architecture used was the Swin UNETR (Figure \ref{fig:swinunetr_architecture}), a U-Net-like network in which the convolutional encoder is replaced by Swin transformer blocks. This change in the encoder makes it possible for the architecture to be capable of capturing long-range information, as opposed to the usual fully convolutional networks \cite{hatamizadeh2021swin}. This network's transformer also allows the use of high-resolution images, as is the case of the BraTS dataset, by using shifted windows.

To maintain repeatability and consistency, the brain voxels of each scan were normalised using z-score normalisation, before the training step, keeping the background at zero. The remaining pre-processing for the data used in the networks, including regular data augmentation, was handled by the nnUNet pipeline.

\subsection{Post-processing and Ensemble strategy}
To ensure maximum success each model was trained with a 5-fold cross-validation resampling method, the ensemble was then averaged together to achieve a final segmentation prediction for each validation image.

Some post-processing was also applied to the predicted images, based on a region-based threshold, to remove some small tumours that could be detected but were not actually tumours. For this, several values were tested for each of the four regions of interest of Task 1 (WT, TC, ET and RC) and for the GTV of Task 3, removing regions marked as tumours with a size below the established threshold.

\section{Results}

We refer to our solutions using the following abbreviations:
\begin{itemize}
    \item \textbf{Task 1 - Adult Glioma Post Treatment (table \ref{tab:task1_val}):}
    \begin{itemize}
        \item \textbf{$RN_g$}: nnUNet (3D full resolution) trained only with the real dataset.
        \item \textbf{$RS_g$}: Swin UNETR trained only with the real dataset.
        \item \textbf{$RM_g$}: MedNeXt trained only with the real dataset.
        \item \textbf{$rGN_g$}: nnUNet (3D full resolution) trained with both real and synthetic data.
        \item \textbf{$rGS_g$}: Swin UNETR trained with both real and synthetic data.
        \item \textbf{$rGM_g$}: MedNeXt trained with both real and synthetic data.
    \end{itemize}
    \item \textbf{Task 3 - Meningioma Radiotherapy (table \ref{tab:task3_val}):}
    \begin{itemize}
        \item \textbf{$RN_m$}: nnUNet (3D full resolution) trained only with the real dataset.
        \item \textbf{$RM_m$}: MedNeXt trained only with the real dataset.
        \item \textbf{$rGN_m$}: nnUNet (3D full resolution) trained with both real and synthetic data.
        \item \textbf{$rGM_m$}: MedNeXt trained with both real and synthetic data.
    \end{itemize}
    
\end{itemize}

\begin{table}[]
\centering
\caption{Validation set results for Task 1 computed by the validation platform. The "All" is defined as $rGN_{g}RN_{g}rGS_{g}RS_{g}rGM_{g}RM_{g}$. The "All real" is composed by all models only trained with real data. The "All fake" is composed by all models trained with real and synthetic data.}
\label{tab:task1_val}
\begin{tabular}{|c|c|c|c|c|c|c|c|c|}
\hline
\textbf{Ensemble} & \begin{tabular}[c]{@{}c@{}}\textbf{Thresholding}\\ (WT, TC, ET, RC)\end{tabular} & \textbf{Metric} & \textbf{ET} & \textbf{NETC} & \textbf{RC} & \textbf{SNFH} & \textbf{TC} & \textbf{WT} \\ \hline
\multirow{2}{*}{All} & \multirow{2}{*}{50, 0, 0, 50} & DSC & \textbf{0.7557} & \textbf{0.7868} & \textbf{0.7053} & 0.8704 & \textbf{0.7500} & 0.8734 \\ \cline{3-9} 
 &  & HD95 & \textbf{34.59} & \textbf{39.81} & 56.97 & 25.05 & \textbf{35.61} & 26.32 \\ \hline
\multirow{2}{*}{All} & \multirow{2}{*}{50, 50, 50, 50} & DSC & 0.7435 & \textbf{0.7868} & \textbf{0.7053} & 0.8703 & 0.7228 & 0.8737 \\ \cline{3-9} 
&  & HD95 & 43.98 & 39.82 & 56.97 & 25.06 & 49.92 & 26.23 \\ \hline
\multirow{2}{*}{All} & \multirow{2}{*}{0, 0, 0, 0} & DSC & \textbf{0.7557} & \textbf{0.7868} & 0.6927 & 0.8678 & \textbf{0.7500} & 0.8727 \\ \cline{3-9} 
&  & HD95 & \textbf{34.59} & \textbf{39.81} & 59.82 & 25.91 & \textbf{35.61} & 26.28 \\ \hline
\multirow{2}{*}{All real} & \multirow{2}{*}{0, 0, 0, 0}& DSC & 0.7506 & 0.7715 & 0.7050 & 0.8715 & 0.7366 & \textbf{0.8759} \\ \cline{3-9} 
&  & HD95 & 35.80 & 48.44 & \textbf{55.92} & 25.51 & 38.52 & 25.65 \\ \hline
\multirow{2}{*}{All fake} & \multirow{2}{*}{0, 0, 0, 0}& DSC & 0.7416 & 0.7764 & 0.6826 & 0.8278 & 0.7327 & 0.8381 \\ \cline{3-9} 
&  & HD95 & 42.78 & 44.40 & 64.43 & 34.97 & 43.33 & 33.98 \\ \hline
\multirow{2}{*}{$rGN_{g}RN_{g}$} & \multirow{2}{*}{0, 0, 0, 0} & DSC & 0.7513 & 0.7151 & 0.6907 & 0.8537 & 0.7444 & 0.8661 \\ \cline{3-9} 
&  & HD95 & 35.21 & 59.10 & 61.62 & 28.64 & 38.92 & 27.37 \\ \hline
\multirow{2}{*}{$rGS_{g}RS_{g}$} & \multirow{2}{*}{0, 0, 0, 0} & DSC & 0.7372 & 0.6768 & 0.6850 & 0.8458 & 0.7250 & 0.8509 \\ \cline{3-9} 
&  & HD95 & 40.53 & 66.09 & 57.67 & 30.53 & 43.15 & 31.22 \\ \hline
\multirow{2}{*}{$rGM_{g}RM_{g}$} & \multirow{2}{*}{0, 0, 0, 0} & DSC & 0.7450 & 0.7626 & 0.6874 & 0.8687 & 0.7220 & 0.8724 \\ \cline{3-9} 
&  & HD95 & 40.53 & 55.03 & 63.52 & 26.13 & 45.98 & 24.47 \\ \hline
\multirow{2}{*}{$RN_{g}$} & \multirow{2}{*}{0, 0, 0, 0} & DSC & 0.7261 & 0.7181 & 0.6906 & 0.8699 & 0.7257 & 0.8714 \\ \cline{3-9} 
&  & HD95 & 44.55 & 58.55 & 59.22 & \textbf{22.95} & 42.82 & 25.21 \\ \hline
\multirow{2}{*}{$rGN_{g}$} & \multirow{2}{*}{0, 0, 0, 0} & DSC & 0.7331 & 0.6865 & 0.6912 & 0.8379 & 0.7271 & 0.8531 \\ \cline{3-9} 
&  & HD95 & 42.62 & 67.55 & 62.67 & 32.53 & 42.73 & 30.84 \\ \hline
\multirow{2}{*}{$RS_{g}$} & \multirow{2}{*}{0, 0, 0, 0} & DSC & 0.7097 & 0.7203 & 0.6901 & 0.8391 & 0.7127 & 0.8478 \\ \cline{3-9} 
&  & HD95 & 49.53 & 58.87 & 56.02 & 32.19 & 46.06 & 32.19 \\ \hline
\multirow{2}{*}{$rGS_{g}$} & \multirow{2}{*}{0, 0, 0, 0} & DSC & 0.7212 & 0.6779 & 0.6547 & 0.8221 & 0.7123 & 0.8317 \\ \cline{3-9} 
&  & HD95 & 46.24 & 66.25 & 70.09 & 36.60 & 48.22 & 37.40 \\ \hline
\multirow{2}{*}{$RM_{g}$} & \multirow{2}{*}{0, 0, 0, 0} & DSC & 0.7542 & 0.7632 & 0.6849 & \textbf{0.8737} & 0.7234 & 0.8751 \\ \cline{3-9} 
&  & HD95 & 37.76 & 56.76 & 63.65 & 25.18 & 45.64 & \textbf{24.39} \\ \hline
\multirow{2}{*}{$rGM_{g}$} & \multirow{2}{*}{0, 0, 0, 0} & DSC & 0.7358 & 0.7725 & 0.6817 & 0.8635 & 0.7103 & 0.8670 \\ \cline{3-9} 
&  & HD95 & 41.91 & 52.28 & 64.07 & 27.00 & 48.25 & 25.65 \\ \hline
\end{tabular}
\end{table}

\begin{table}[]
\centering
\caption{Validation set results for Task 3 computed by the validation platform.}
\label{tab:task3_val}
\begin{tabular}{|c|c|c|c|c|c|c|}
\hline
\textbf{Solution} & \textbf{Threshold} & \textbf{DSC} & \textbf{HD95} & \textbf{FP} & \textbf{FN} \\ \hline
$RN_mrGN_mRM_mrGM_m$ & \multicolumn{1}{c|}{0}& \multicolumn{1}{c|}{0.79789} & \multicolumn{1}{c|}{50.6550}& \multicolumn{1}{c|}{\textbf{0}} & \multicolumn{1}{c|}{11}\\
$RN_mrGN_mRM_mrGM_m$ & \multicolumn{1}{c|}{300}& \multicolumn{1}{c|}{0.79597} & \multicolumn{1}{c|}{52.0605}& \multicolumn{1}{c|}{\textbf{0}} & \multicolumn{1}{c|}{12}\\
$RN_m$ & \multicolumn{1}{c|}{0}& \multicolumn{1}{c|}{0.78907} & \multicolumn{1}{c|}{51.6856} & \multicolumn{1}{c|}{4} & \multicolumn{1}{c|}{12}\\ 
$RN_m$ & \multicolumn{1}{c|}{300}& \multicolumn{1}{c|}{0.78914} & \multicolumn{1}{c|}{51.6340}& \multicolumn{1}{c|}{1} & \multicolumn{1}{c|}{12}\\
$rGN_m$ & \multicolumn{1}{c|}{0}& \multicolumn{1}{c|}{0.77100} & \multicolumn{1}{c|}{64.5203}& \multicolumn{1}{c|}{4} & \multicolumn{1}{c|}{13}\\
$RM_m$ & \multicolumn{1}{c|}{0}& \multicolumn{1}{c|}{0.80322} & \multicolumn{1}{c|}{41.6789}& \multicolumn{1}{c|}{6} & \multicolumn{1}{c|}{9}\\
$rGM_m$ & \multicolumn{1}{c|}{0}& \multicolumn{1}{c|}{\textbf{0.82144}} & \multicolumn{1}{c|}{\textbf{24.6422}}& \multicolumn{1}{c|}{8} & \multicolumn{1}{c|}{\textbf{6}}\\
\hline
\end{tabular}
\end{table}

\begin{table}[h!]
\centering
\caption{Test set results for Task 1 computed by the challenge organisers.}
\label{tab:task1_test}
\begin{tabular}{|c|c|c|c|c|}
\hline
\textbf{Region} & \textbf{DSC (Mean)} & \textbf{DSC (Std)} & \textbf{HD95 (Mean)} & \textbf{HD95 (Std)} \\
\hline
ET   & 0.7900 & 0.2958 & 35.63 & 89.86 \\
NETC & 0.8076 & 0.3207 & 30.35 & 90.15 \\
RC   & 0.7760 & 0.3263 & 44.58 & 106.78 \\
SNFH & 0.8926 & 0.1591 & 16.87 & 47.48 \\
TC   & 0.7874 & 0.2883 & 38.19 & 90.61 \\
WT   & 0.8938 & 0.1634 & 17.95 & 49.36 \\
\hline
\end{tabular}
\end{table}

\begin{table}[h!]
\centering
\caption{Test set results for Task 3 computed by the challenge organisers.}
\label{tab:task3_test}
\begin{tabular}{|c|c|c|c|c|}
\hline
\textbf{Region} & \textbf{DSC (Mean)} & \textbf{DSC (Std)} & \textbf{HD95 (Mean)} & \textbf{HD95 (Std)} \\
\hline
GTV   & 0.801 & 0.221 & 38.26 & 139.45 \\
\hline
\end{tabular}
\end{table}

\section{Discussion and Conclusion}
The results of the validation set for tasks 1 and 3 are presented in tables \ref{tab:task1_val} and \ref{tab:task3_val}. The test results of task 1 and 3 provided by the organisers are presented in tables \ref{tab:task1_test} and \ref{tab:task3_test}.

Regarding task 1, from table \ref{tab:task1_val} it can be seen that it is still room for improvement, specially of the RC region which had the best DSC around 0.7. Similarly to BraTs 2023 Task 1 solution \cite{ferreira2024we}, the ensemble of several models shows a very good performance compared to each individual model, as the number of false positives (FP) is reduced without significantly increasing the number of FN. The models trained with synthetic data, when tested individually, have a lower number of FN, however a considerably higher number of FP which heavily penalises the results. The random generation of synthetic tumours anywhere in the brain might therefore potentially make the training process for the segmentation very unstable and the model very prone to FP, as too many non-existent tumours are predicted, specially the Swin UNETR. The models trained with MexNeXt have less FP than the others, but more FN. Therefore, it was determined that the ensemble of all models would have the greatest potential for the test phase as it was balanced in terms of FP and FN. We decided on a threshold of 50 for WT and 50 for RC in the post-processing step. Using larger thresholds produced worst results, meaning that our final solution did not produce many small FP. With this solution, we have achieved the DSC and HD95 specified in the first entry of the table \ref{tab:task1_val} in the validation phase and the best ranking in the testing phase. The DSC and HD95 of the testing phase are presented in table \ref{tab:task1_test}.

For task 3 a semi-lesion-wise metric was used to evaluate the predictions, i.e., the evaluation metric does not penalise the FP as these can be easily identified and removed by a treating radiation oncologist. However, the FN are still penalised. In this case, the use of a threshold does not improve much the results, as can be seen in Table \ref{tab:task3_val}. It can even produce worse results by removing predictions that are true positives, creating FN. Unlike in Task 1, there is no good balance between the number of FP and FN. The use synthetic data to train the MedNeXt framework had a positive effect on the results. The number of FN is the lowest among all the solutions, which had a direct effect on the lesion-wise DSC. As mentioned in the section "\ref{DataAugmentation} Data augmentation", the pipeline for generating synthetic data was not fully adapted to this dataset. However, the randomness of the locations where the synthetic tumour was located made the network more sensible and robust, avoiding FN. The most successful model tested was then the MexNeXt trained with real and synthetic data and no threshold, attaining a DSC of 0.8214 and HD95 of 24.64 in the validation phase. Furthermore, this version of the model was able to achieve the lowest number of FN of the ones tested. Therefore, this was the solution submitted to the testing phase, archiving the results presented in table \ref{tab:task3_test}.

For future work, it would be important to further improve the GliGAN pipeline to select better locations for the placement of the synthetic tumour and to generate more realistic cases. Statistical similarity assessments should also be used to evaluate the quality of the synthetic tumours, rather than just a visual inspection.

\begin{credits}
\subsubsection{\ackname} André Ferreira thanks the Fundação para a Ciência e Tecnologia (FCT) Portugal for the grant 2022.11928.BD. Tiago Jesus thanks the same institution for the grant 2021.05068.BD. This work was supported by FCT within the R\&D Units Project Scope: UIDB/00319/2020. It was also supported by the Advanced Research Opportunities Program (AROP) of RWTH Aachen University and Clinician Scientist Program of the Faculty of Medicine RWTH Aachen University, also some of the computations were performed with computing resources granted by RWTH Aachen University under project 
rwth1484. This work received funding from enFaced (FWF KLI 678), enFaced 2.0 (FWF KLI 1044) and KITE (Plattform für KI-Translation Essen) from the REACT-EU initiative (EFRE-0801977, \url{https://kite.ikim.nrw/}).Data used in this publication were obtained as part of the Challenge project through Synapse ID (syn51156910).

\subsubsection{\discintname}
The authors have no competing interests to declare that are
relevant to the content of this article.
\end{credits}
%
%
%

\bibliographystyle{splncs04}
\bibliography{bib}
\end{document}